\pdfoutput=1
\documentclass[11pt]{article}

\usepackage[final]{acl}

\usepackage{times}
\usepackage{latexsym}

\usepackage[T1]{fontenc}
\usepackage[utf8]{inputenc}

\usepackage{microtype}

\usepackage{inconsolata}

\usepackage{graphicx}

\usepackage{amsmath}
\usepackage{amssymb}
\usepackage{xcolor}
\usepackage{multirow} 
\usepackage{booktabs}
\usepackage{multicol}
\usepackage{colortbl}
\usepackage{tikz,pgfplots}
\usepackage{tcolorbox}
\usepackage{algorithm}
\usepackage{algpseudocode}
\usepackage{hyperref}
\usepackage{silence}
\usepackage{fontawesome5}
\usepackage{MnSymbol}
\usepackage{amssymb}
\usepackage{pifont}
\newcommand{\xmark}{\ding{55}}%

\pgfplotsset{compat=1.18}

\title{Image Embedding Sampling Method for Diverse Captioning}

\author{
      Sania Waheed\footnotemark[1]\footnotemark[2] \\
      University of Southhampton \\
      \texttt{sw1m24@soton.ac.uk} \\\And
      Na Min An\footnotemark[1] \\
      KAIST AI \\
      \texttt{naminan@kaist.ac.kr}}

\begin{document}
\maketitle

\begin{abstract}
Image Captioning for state-of-the-art VLMs has significantly improved over time; however, this comes at the cost of increased computational complexity, making them less accessible for resource-constrained applications such as mobile devices and assistive technologies. Alternatively, comparably smaller VLMs prioritize high-level scene descriptions, overlooking finer details that contribute to a richer understanding of an image. In this paper, we introduce a training-free framework that enhances caption diversity and informativeness by explicitly attending to distinct image regions using a comparably small VLM, BLIP, as the backbone. Our approach leverages structured segmentation to produce hierarchical representations that capture both global and localized semantics. Without requiring additional model training, we demonstrate that our method allows smaller VLMs to achieve performance comparable to larger models in terms of image-caption alignment, semantic integrity, and diversity. We evaluate our framework on MSCOCO, Flickr30k, and Nocaps test datasets, achieving a Div-2 score of 0.735, 0.750, and 0.748 for each dataset, respectively, while maintaining strong image-caption relevancy and semantic integrity with the human-annotated captions. Our code is available at \url{https://github.com/xfactlab/HBoP}.
\end{abstract}

\footnotetext[1]{Equal contribution.}
\footnotetext[2]{Work done while at KAIST.}

\section{Introduction}
\label{ch:intro}

\begin{figure*}[t!]
\centering
\includegraphics[width=\textwidth]{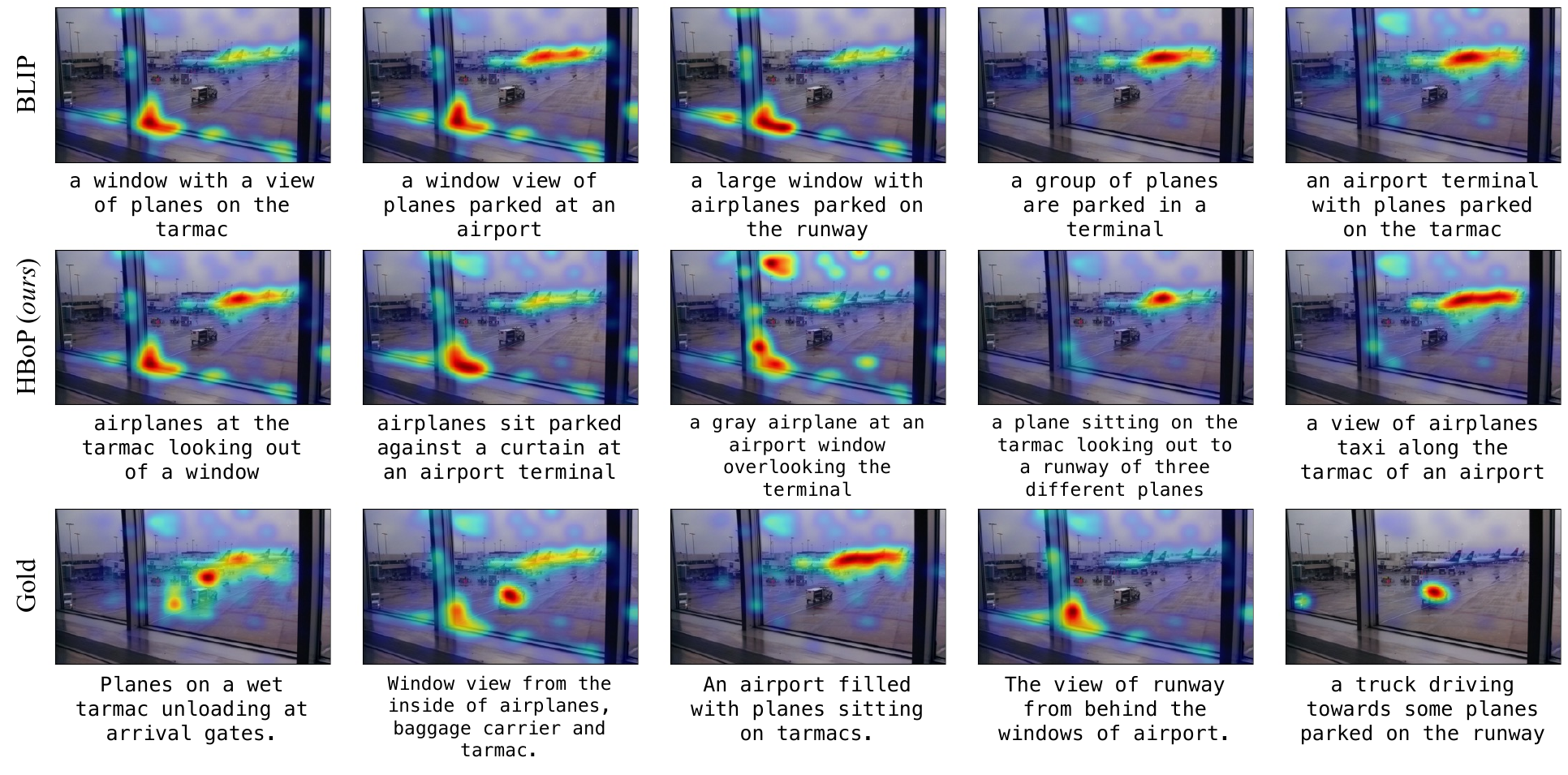}
\caption{Comparison of captions generated by BLIP, HBoP, and human annotations. The images are overlaid with GradCAM heatmaps to highlight the regions focused on by the pretrained image-text matching model \cite{li2022blip}. HBoP captions exhibit greater diversity compared to BLIP captions and are closer to human-annotated gold captions.}
\label{fig:BLIP-2vsHBoP}
\end{figure*}

Visual-Language Models (VLMs) have seen rapid advancements in image captioning, benefiting from increasingly sophisticated architectures and larger training datasets \cite{alayrac2022flamingo, li2022blipbootstrappinglanguageimagepretraining, radford2021learningtransferablevisualmodels, wang2022imageforeignlanguagebeit}. State-of-the-art large-scale models generate highly detailed and diverse captions, yet their extensive computational requirements can be prohibitive in resource-constrained settings. Conversely, smaller VLMs, while more efficient, often prioritize dominant visual elements and overlook fine-grained details, resulting in captions that lack the depth and specificity seen in human-generated captions \cite{aneja2019sequentiallatentspacesmodeling, bianco2023improvingimagecaptioningdescriptiveness, chen2023learningdistinctrepresentativestyles, yuksekgonul2022and,an2025can}.

Inspired by previous work \cite{ji2021step, Shao2023ICCV, shukor2022efficient} that demonstrates the advantages of hierarchical approaches in image understanding, our method leverages structured segmentation to capture both global and regional aspects of an image. We sample segmentation-driven embeddings from the last layer of the visual encoder, where self-attention mechanisms have already propagated information across the entire image. This allows the model to explicitly attend to distinct local image regions while preserving contextual relationships, generating captions at multiple levels of granularity. This approach offers an efficient alternative to enhancing caption diversity in smaller VLMs without LLMs, achieving performance comparable to larger LLM-based models in terms of caption diversity and image-caption alignment.

We validate our approach, namely, \textbf{HBoP} - \textbf{H}ierarchical \textbf{B}ags \textbf{o}f \textbf{P}hrases, by evaluating generated captions for MSCOCO \cite{lin2014microsoft}, Flickr30k \cite{young2014image}, and Nocaps \cite{agrawal2019nocaps} datasets on conventional diversity metrics such as mBLEU-4, n-gram diversity \cite{aneja2019sequential}, and newly presented pairwise cosine distance (PCD). Our findings show that structured caption generation effectively improves diversity while maintaining relevancy with images and human-generated captions (\textit{compare} BLIP \cite{li2022blip}, HBoP, and gold captions in Figure~\ref{fig:BLIP-2vsHBoP}). 

\section{Related Works}
\label{ch:relatedwork}
Vision-language models have shown strong performance in multimodal tasks, with caption generation as a key benchmark. Models like CLIP \cite{radford2021learningtransferablevisualmodels}, Flamingo \cite{alayrac2022flamingo}, and BLIP-2 \cite{li2023blip} use contrastive learning and large-scale pre-training to enhance vision-language alignment. However, they often produce high-level scene descriptions, missing fine-grained details needed for detailed image understanding. Traditional captioning approaches treat images holistically, overlooking hierarchical details \cite{xu2021towards}, unless explicitly trained for diversity, as in ModeCap \cite{chen2022learning} and Seq-CVAE \cite{aneja2019sequential}.

Inspired by hierarchical representation techniques \cite{ji2021step, Shao2023ICCV,shukor2022efficient}, our approach samples latent image embeddings from structured segmentation to generate multi-level captions. This aligns with recent region-based methods using SAM \cite{shlapentokhrothman2024regionbasedrepresentationsrevisited} and studies on caption quality focused on informational sufficiency, minimal redundancy, and human comprehensibility \cite{chen2024makesgoodimagecaptions}. Our evaluation metrics reflect these aspects: CLIP score for informational sufficiency, mBLEU and Div-2 for redundancy, and SBERT for comprehensibility.

\begin{figure*}[t!]
\centering
\includegraphics[width=\textwidth]{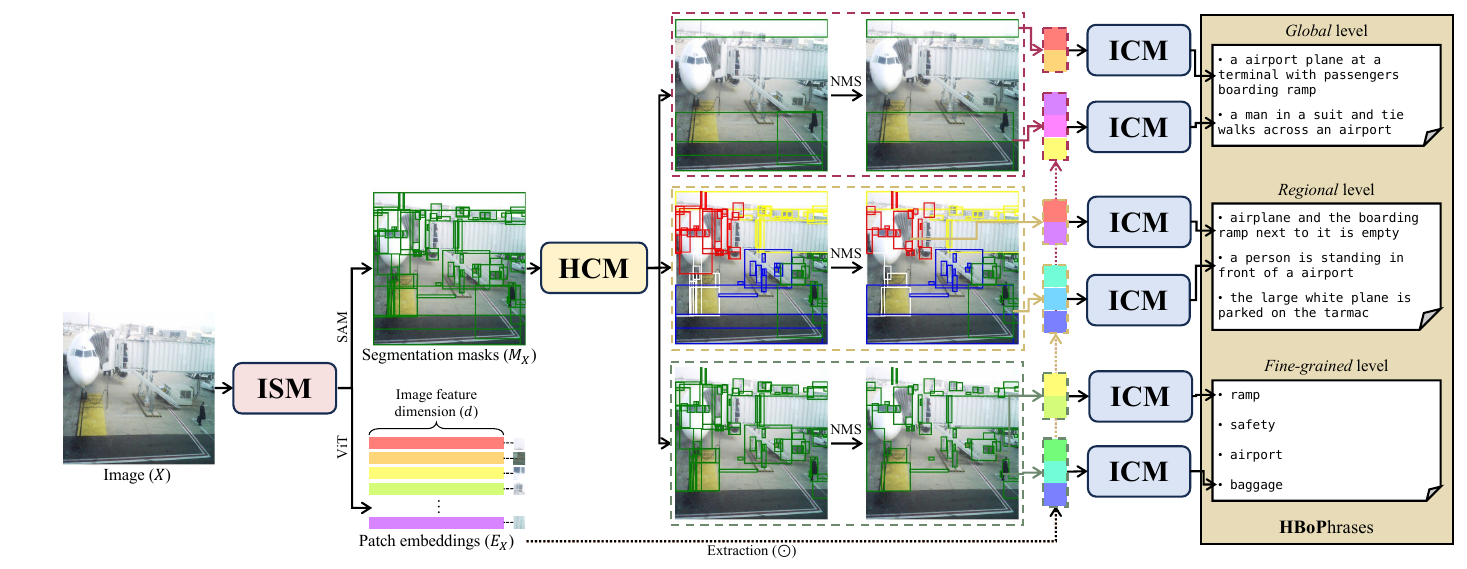}
\caption{The proposed HBoP framework consists of three components: (1) Image Segmentation Module (ISM), (2) Hierarchical Composition Module (HCM), and (3) Image Captioning Module (ICM). HBoP controls caption granularity by selecting meaningful patch embeddings of varying sizes from the segmentation model. For instance, for the regional level caption generation, although the selected regions correspond to specific visual entities (e.g., the “airplane” in red boxes or the “person” in green boxes), the generated captions go beyond these isolated regions to describe broader scene elements, such as the “boarding ramp” or the “airport.”}
\label{fig:HBoP}
\end{figure*}

\section{Methodology}
\label{ch:ourmethod}

In this section, we introduce our proposed framework, HBoP (depicted in Fig~\ref{fig:HBoP}), a modular architecture that uses pre-trained segmentation and captioning models. We show that HBoP ensures multiple levels of captions (\textit{i.e.}, global, regional, fine-grained) by inducing a hierarchical structure for image understanding.

\subsection{Image Segmentation Module (ISM)}\label{method:ims}
The first component of HBoP, ISM, selects patch embeddings ($E_{X}$) corresponding to image regions ($X = (X_1, X_2, ..., X_n)$) from the original image embeddings extracted using a Vision Transformer (ViT) \cite{dosovitskiy2020image} encoder. These regions are selected based on segmentation masks produced by a segmentation model. In our implementation, we use the Segment Anything Model (SAM) \footnote{While we use SAM in our experiments, the HBoP framework is flexible and compatible with any segmentation model that can provide region masks. Additionally, several prior works \cite{suo-etal-2023-text, Yu_2023_CVPR, wang2025iterprime} have adopted training-free methods that incorporate unsupervised segmentation models for similar purposes. Additionally, a recent efficient implementation of SAM achieves up to 50× higher run-time speed, helping address concerns around computational overhead \cite{zhao2023fast}.} \cite{kirillov2023segany} due to its strong segmentation performance across diverse benchmarks. For a set of $p$ segmentation masks in the image, the resulting masks for the selected image regions would be: $M_{X} = \{\,M_{X_1}, M_{X_2}, ..., M_{X_p}\,\} = \text{SAM}(X), X \in \mathbb{R}^{H \times W \times C}$, where $H$, $W$, and $C$ represent the height, width, and channels of $X$.

\subsection{Hierarchical Composition Module (HCM)}\label{method:hcm}

The second component, HCM, is a key component that can control the level of captions. Specifically, we present three types of captions that can be derived using HCM. 
\paragraph{\textit{Global/Fine-grained level} captions} The global segmentation masks ($M_{G}$) are selected by choosing the top-$k$ (5 in our case) largest segmentation masks from $M_X$ after applying non-maximum suppression (NMS)\footnote{NMS introduces a hyperparameter (IoU threshold), which is set to 0.1 in this case, to aggressively filter overlapping masks. While this step introduces a minor deviation from end-to-end processing, the hyperparameter is intuitive and fixed, requiring minimal tuning.}\cite{hosang2017learning}:
\begin{align*}
    M_G &= \{M_{g_1}, M_{g_2}, ..., M_{n_g} \}, \\
    M_{g_i} &= \text{NMS}(\text{Top-}k (M_X)), \quad i=1, ..., n_g
\end{align*}
NMS removes multiple segmentation masks with overlapping, similar contexts using the Intersection over Union (IoU) and predicted confidence from SAM. The remaining masks, after applying NMS, can also be used to generate fine-grained captions (discussed in Appendix~\ref{app:fine}): 
\begin{align*}
    M_F &= \{M_{f_1}, M_{f_2}, ..., M_{n_f} \}, \\
    M_{f_i} &= \text{NMS}(M_X) \setminus M_G, \quad i=1, ..., n_f
\end{align*}

\paragraph{\textit{Regional level} captions} To create regional-level segmentation masks, $M_{R}$, we use $K$-means clustering to partition all the segmentation masks ($M_X$) and apply NMS to each cluster individually:
\begin{align*}
     M_R &= \{M_{r_1}, M_{r_2}, ..., M_{K} \}, \\
     M_{r_i} &= \text{NMS}(\text{K-means}(M_X)), i=1, ..., K
\end{align*}

The hierarchical segmentation masks ($M_{g}$, $M_{r}$ and $M_{f}$) are used to extract relevant patch embeddings, $E_{g}$, $E_{r}$ and $E_{f}$ using $E_{X}$ from the first stage. We extract ($\odot$) the corresponding embeddings by concatenating the extracted patch embeddings of different levels (see Appendix~\ref{app:impdetails}). Thus, the final selected image embeddings can be categorized as:
%\vspace{-3mm}
\begin{align*}
    E_{G} &= \{\,E_{g_1}, E_{g_2}, ..., E_{g_{n_g}}\,\}, E_{g_i} = E_X \odot M_{g_i}\\
    E_{R} &= \{\,E_{r_1}, E_{r_2}, ..., E_{K}\,\}, E_{r_i} = E_X \odot M_{r_i}\\
    E_{F} &= \{\,E_{f_1}, E_{f_2}, ..., E_{n_f}\,\}, E_{f_i} = E_X \odot M_{f_i}
\end{align*}

\begin{table*}[t!]
\begin{center}
\resizebox{\textwidth}{!}{%
\begin{tabular}{l|c|c|cc|ccc|cc|ccc}
\toprule
\multicolumn{1}{l|}{} & & & \multicolumn{5}{c|}{MSCOCO (5k test set)}  & \multicolumn{5}{c}{Flickr30K (1k test set)}  \\
& \multirow{2}{*}{LLM} & \# of & \multicolumn{2}{c|}{Relevancy} & \multicolumn{3}{c|}{Diversity}  & \multicolumn{2}{c|}{Relevancy} & \multicolumn{3}{c}{Diversity}  \\

&  & Param & SBERT $\uparrow$  & CLIP-S $\uparrow$ & PCD & mBLEU-4 $\downarrow$ & Div-2 $\uparrow$ & SBERT $\uparrow$ & CLIP-S $\uparrow$ & PCD & mBLEU-4 $\downarrow$ & Div-2 $\uparrow$ \\

\midrule
\textit{Random} & - & - & - &  17.77 &  0.963 & 0.001  & 0.868 & - &  17.54 & 0.962 &  0.003 &  0.860  \\

\midrule
BLIP ($-$NS) & \xmark & 446M & \cellcolor[HTML]{ADD8E6} 56.00 & \cellcolor[HTML]{FFCCCC} 29.98 & \cellcolor[HTML]{ADD8E6} 0.600 & \cellcolor[HTML]{FFCCCC} 1.000 & \cellcolor[HTML]{ADD8E6} 0.179 & \cellcolor[HTML]{FFCCCC} 55.78 & \cellcolor[HTML]{FFCCCC} 28.58 &  \cellcolor[HTML]{ADD8E6} 0.600 & \cellcolor[HTML]{FFCCCC} 1.000 & \cellcolor[HTML]{ADD8E6} 0.179  \\

BLIP ($+$NS) & \xmark &  446M  & \cellcolor[HTML]{FFCCCC} 57.23 & \cellcolor[HTML]{FFCCCC} 30.33 & \cellcolor[HTML]{ADD8E6}  0.668 & \cellcolor[HTML]{FFCCCC} 0.658 &  \cellcolor[HTML]{ADD8E6} 0.387 & \cellcolor[HTML]{ADD8E6} 46.99 & \cellcolor[HTML]{FFCCCC} 29.56 &  \cellcolor[HTML]{ADD8E6} 0.690 & \cellcolor[HTML]{FFCCCC} 0.664 & \cellcolor[HTML]{ADD8E6} 0.384 \\

Seq-CVAE & \xmark & - & - & - & - & \cellcolor[HTML]{FFCCCC} 0.640 & \cellcolor[HTML]{ADD8E6} 0.480 & - & - & - & - & -  \\

ModeCap & \xmark & - & - &\cellcolor[HTML]{FFCCCC} 29.35 & \cellcolor[HTML]{ADD8E6} 0.714 & \cellcolor[HTML]{FFCCCC} 0.281 & \cellcolor[HTML]{ADD8E6} 0.594 & - & - & - & - & -\\

\midrule

BLIP-2 & $\checkmark$ & 3.9B & 65.47 & 30.66 & 0.651  & 0.712  & 0.345 & 57.81 & 30.37 & 0.667 & 0.732  & 0.336   \\

Honeybee  & $\checkmark$ &  7B &  53.55 &  28.21 & 0.792 &  0.062  &  0.716 &  47.41 &  27.65 & 0.827 &   0.057 &  0.732  \\

Honeybee  & $\checkmark$ &  13B & 55.11 
 & 27.41 & - & 0.014  & 0.872 & 50.41 &  27.27 & - & 0.013 & 0.875  \\

LLaVA-1.5  & $\checkmark$ &  13B &  59.61 
 &  30.08 & - &  0.180  &  0.658 &  54.74 &  29.54 & - &  0.176 &  0.680  \\

LLaVA-1.6  & $\checkmark$ &  7B &  55.99 &  29.36 & - & 0.046  &  0.787 &  51.00  & 27.46 & - &  0.028 &  0.809  \\

\textit{Gold} & - & - & - &  30.33 & 0.753  &  0.043  &  0.748 & - &  30.87 & 0.776 &  0.049  &  0.760 \\

\midrule
HBoP (\textit{ours}) & \xmark & 1B & 56.30 & 29.12 & 0.772 & 0.049 & 0.735 & 54.00 & 28.46 & 0.815 & 0.042 & 0.750  \\

\multicolumn{3}{c|}{HBoP \textit{Ranking}} & 4/8 & 8/11 & 1/7 & 5/12 & 5/12 & 4/8 & 6/10 & 1/6 & 4/10 & 5/10  \\
\bottomrule
\end{tabular}
}
\caption{Relevance and diversity scores across different models on the MSCOCO and Flickr30K datasets. HBoP achieves stronger diversity with higher Div-2 and PCD scores and a lower mBLEU-4 score compared to smaller VLMs and models trained to enhance diversity, while maintaining comparable relevance scores (SBERT and CLIP-S). Additionally, HBoP demonstrates competitive performance relative to much larger LLM-based VLMs. Cell colors indicate relative comparison to HBoP, with red showing higher values and blue showing lower values. Arrows next to each metric denote whether a higher (↑) or lower (↓) value indicates better performance. }
\label{table:Div}
\end{center}
\end{table*}

\vspace{-2.17mm}
\subsection{Image Captioning Module (ICM)}\label{method:icm}
To generate captions for different levels of image embeddings, we use BLIP fine-tuned on image captioning \cite{li2022blip} with the stochastic sampling method, following the same procedure as \cite{tiong2022plug}. The caption generation process is repeated for $n_g$, $n_r$, and $n_f$ patch embeddings corresponding to the number of selected hierarchical masks. Since the patch embedding size may vary due to the different mask sizes, we use zero padding before using the captioning module. Our final HBoP captions would be:
%\vspace{-3mm}
\begin{align*}
    \text{HBoP}_{G} &= \{\,s_{g_1}, ..., s_{n_g}\,\}, s_{g_i} = \text{BLIP}(E_{g_i})\\
    \text{HBoP}_{R} &= \{\,s_{r_1}, ..., s_{K}\,\}, s_{r_i} = \text{BLIP}(E_{r_i})\\
    \text{HBoP}_{F} &= \{\,s_{f_1}, ..., s_{n_f}\,\}, s_{f_i} = \text{BLIP}(E_{f_i})
\end{align*}
%\vspace{-3mm}
\section{Results}
\noindent

\noindent

\textbf{HBoP achieves the best diversity scores while maintaining relevance among smaller VLMs.}
We evaluate the diversity and relevance of captions generated by different models in Table~\ref{table:Div}, using five captions per image. For HBoP, two global and three regional captions are sampled\footnote{Fine-grained captions are excluded from this evaluation because they function more as image tags than full descriptive captions.}. Although HBoP increases the parameter count relative to BLIP, it remains significantly smaller than VLMs with LLMs, achieving a strong trade-off between diversity and model size. HBoP consistently achieves diversity scores closest to the gold-standard captions among smaller models, as measured by PCD (see Appendix~\ref{app:exp}), mBLEU-4, and Div-2~\cite{aneja2019sequential}. Specifically, it reduces mBLEU-4 by over 60\% and improves Div-2 by more than 30\% compared to BLIP (NS) while maintaining comparable relevancy scores. We also compare our embedding-sampling approach to a baseline where segmented regions are directly cropped and captioned; while cropping improves diversity, it reduces relevance due to loss of global context. Full results are provided in Appendix~\ref{app:exp}.

Compared to baselines such as BLIP~\cite{li2022blip}, Seq-CVAE~\cite{aneja2019sequential}, and ModeCap\footnote{The dataset annotations and features necessary to train ModeCap are exclusively available for the MSCOCO dataset, making it difficult to replicate the experiments for fair comparison on the NoCaps and Flickr30k datasets.}\cite{chen2022learning}, HBoP achieves the lowest mBLEU-4 and highest Div-2 scores. Notably, it even outperforms larger models like BLIP-2, Honeybee-7B\cite{cha2023honeybee}, and LLaVA-1.5~\cite{liu2023llava} in several diversity metrics, despite using ~4× to 13× fewer parameters. This highlights HBoP’s effectiveness as a lightweight alternative for generating diverse captions without the overhead of large-scale models.

HBoP maintains strong similarity between generated captions and reference texts and image-text alignment, as measured by SBERT and CLIP-Score, respectively. We use SBERT similarity as a practical proxy for human evaluation, as the reference captions are human-written; thus, high SBERT scores indicate strong semantic alignment and provide an approximate measure of caption quality from a human perspective\footnote{These high SBERT scores also imply that our framework is not negatively affected by the errors or failure cases of SAM. We also provide the performance and efficiency results of FastSAM \cite{zhao2023fast} in Appendix~\ref{app:fastsam}.}. HBoP achieves scores comparable to BLIP, BLIP-NS, and LLaVA, while outperforming HoneyBee. Although BLIP-2 scores the highest, HBoP demonstrates a strong balance between relevance and diversity. Further semantic integrity evaluations are detailed in Appendix~\ref{app:semantic}.

\section{Conclusion}
\noindent
We propose HBoP, a hierarchical caption generation framework that leverages a modular architecture combining lightweight pre-trained VLMs and segmentation models to generate semantically meaningful yet diverse captions. Our experimental results demonstrate HBoP’s ability to produce meaningful image embeddings for captioning, achieving performance comparable to larger VLMs and human-generated captions. HBoP sets a solid baseline for future work aiming to extract more relevant knowledge by controlling the intermediate image embeddings.

\section{Limitations}
\noindent
The current implementation of HBoP relies on bounding box approximations of segmentation masks to extract image embeddings. While effective, this may occasionally miss fine-grained or irregularly shaped image details. Exploring the use of full, irregular-shaped segmentation masks for embedding extraction is a promising direction for future work. Another limitation is that our approach primarily enhances factual diversity by focusing on distinct image regions, but it may be less effective for domains required for capturing cultural diversity \cite{bayramli-etal-2025-diffusion}, as such interpretations often rely on external cultural knowledge beyond visual features.

\section{Ethical Statement}
Captions generated with HBoP might inadvertently contain harmful content. However, the final caption outputs mainly depend on the image content and the pretrained image captioning model. Therefore, unless the images themselves are harmful or the pretrained model produces unsafe captions, HBoP captions are expected to pose minimal risk.  

\section{Acknowledgement}
This work was supported by Institute for Information \& communications Technology Planning \& Evaluation (IITP) grant funded by the Korea government(MSIT) (RS-2019-II190075, Artificial Intelligence Graduate School Program (KAIST)).

\bibliography{custom}

\begin{thebibliography}{57}
\providecommand{\natexlab}[1]{#1}

\bibitem[{Agrawal et~al.(2019)Agrawal, Desai, Wang, Chen, Jain, Johnson, Batra, Parikh, Lee, and Anderson}]{agrawal2019nocaps}
Harsh Agrawal, Karan Desai, Yufei Wang, Xinlei Chen, Rishabh Jain, Mark Johnson, Dhruv Batra, Devi Parikh, Stefan Lee, and Peter Anderson. 2019.
\newblock Nocaps: Novel object captioning at scale.
\newblock In \emph{Proceedings of the IEEE/CVF international conference on computer vision}, pages 8948--8957.

\bibitem[{Alayrac et~al.(2022)Alayrac, Donahue, Luc, Miech, Barr, Hasson, Lenc, Mensch, Millican, Reynolds et~al.}]{alayrac2022flamingo}
Jean-Baptiste Alayrac, Jeff Donahue, Pauline Luc, Antoine Miech, Iain Barr, Yana Hasson, Karel Lenc, Arthur Mensch, Katherine Millican, Malcolm Reynolds, et~al. 2022.
\newblock Flamingo: a visual language model for few-shot learning.
\newblock \emph{Advances in Neural Information Processing Systems}, 35:23716--23736.

\bibitem[{An et~al.(2025)An, Kim, Kang, Kim, Shim, and Thorne}]{an2025can}
Na~Min An, Eunki Kim, Wan~Ju Kang, Sangryul Kim, Hyunjung Shim, and James Thorne. 2025.
\newblock Can lvlms and automatic metrics capture underlying preferences of blind and low-vision individuals for navigational aid?
\newblock \emph{arXiv preprint arXiv:2502.14883}.

\bibitem[{An et~al.(2024)An, Waheed, and Thorne}]{an-etal-2024-capturing}
Na~Min An, Sania Waheed, and James Thorne. 2024.
\newblock \href {https://aclanthology.org/2024.findings-eacl.43} {Capturing the relationship between sentence triplets for {LLM} and human-generated texts to enhance sentence embeddings}.
\newblock In \emph{Findings of the Association for Computational Linguistics: EACL 2024}, pages 624--638, St. Julian{'}s, Malta. Association for Computational Linguistics.

\bibitem[{Aneja et~al.(2019{\natexlab{a}})Aneja, Agrawal, Batra, and Schwing}]{aneja2019sequentiallatentspacesmodeling}
Jyoti Aneja, Harsh Agrawal, Dhruv Batra, and Alexander Schwing. 2019{\natexlab{a}}.
\newblock \href {https://arxiv.org/abs/1908.08529} {Sequential latent spaces for modeling the intention during diverse image captioning}.
\newblock \emph{Preprint}, arXiv:1908.08529.

\bibitem[{Aneja et~al.(2019{\natexlab{b}})Aneja, Agrawal, Batra, and Schwing}]{aneja2019sequential}
Jyoti Aneja, Harsh Agrawal, Dhruv Batra, and Alexander Schwing. 2019{\natexlab{b}}.
\newblock Sequential latent spaces for modeling the intention during diverse image captioning.
\newblock In \emph{Proceedings of the IEEE/CVF International Conference on Computer Vision}, pages 4261--4270.

\bibitem[{Banerjee and Lavie(2005)}]{banerjee-lavie-2005-meteor}
Satanjeev Banerjee and Alon Lavie. 2005.
\newblock \href {https://aclanthology.org/W05-0909} {{METEOR}: An automatic metric for {MT} evaluation with improved correlation with human judgments}.
\newblock In \emph{Proceedings of the {ACL} Workshop on Intrinsic and Extrinsic Evaluation Measures for Machine Translation and/or Summarization}, pages 65--72, Ann Arbor, Michigan. Association for Computational Linguistics.

\bibitem[{Bayramli et~al.(2025)Bayramli, Suleymanzade, An, Ahmad, Kim, Park, Thorne, and Oh}]{bayramli-etal-2025-diffusion}
Zahra Bayramli, Ayhan Suleymanzade, Na~Min An, Huzama Ahmad, Eunsu Kim, Junyeong Park, James Thorne, and Alice Oh. 2025.
\newblock \href {https://doi.org/10.18653/v1/2025.acl-long.1503} {Diffusion models through a global lens: Are they culturally inclusive?}
\newblock In \emph{Proceedings of the 63rd Annual Meeting of the Association for Computational Linguistics (Volume 1: Long Papers)}, pages 31137--31155, Vienna, Austria. Association for Computational Linguistics.

\bibitem[{Bianco et~al.(2023)Bianco, Celona, Donzella, and Napoletano}]{bianco2023improvingimagecaptioningdescriptiveness}
Simone Bianco, Luigi Celona, Marco Donzella, and Paolo Napoletano. 2023.
\newblock \href {https://arxiv.org/abs/2306.11593} {Improving image captioning descriptiveness by ranking and llm-based fusion}.
\newblock \emph{Preprint}, arXiv:2306.11593.

\bibitem[{Cha et~al.(2023)Cha, Kang, Mun, and Roh}]{cha2023honeybee}
Junbum Cha, Wooyoung Kang, Jonghwan Mun, and Byungseok Roh. 2023.
\newblock Honeybee: Locality-enhanced projector for multimodal llm.
\newblock \emph{arXiv preprint arXiv:2312.06742}.

\bibitem[{Chen et~al.(2024)Chen, Cahyawijaya, Ishii, Chan, Bang, and Fung}]{chen2024makesgoodimagecaptions}
Delong Chen, Samuel Cahyawijaya, Etsuko Ishii, Ho~Shu Chan, Yejin Bang, and Pascale Fung. 2024.
\newblock \href {https://arxiv.org/abs/2405.00485} {What makes for good image captions?}
\newblock \emph{Preprint}, arXiv:2405.00485.

\bibitem[{Chen et~al.(2022)Chen, Deng, and Wu}]{chen2022learning}
Qi~Chen, Chaorui Deng, and Qi~Wu. 2022.
\newblock Learning distinct and representative modes for image captioning.
\newblock \emph{Advances in Neural Information Processing Systems}, 35:9472--9485.

\bibitem[{Chen et~al.(2023)Chen, Deng, and Wu}]{chen2023learningdistinctrepresentativestyles}
Qi~Chen, Chaorui Deng, and Qi~Wu. 2023.
\newblock \href {https://arxiv.org/abs/2209.08231} {Learning distinct and representative styles for image captioning}.
\newblock \emph{Preprint}, arXiv:2209.08231.

\bibitem[{Chiang and yi~Lee(2023)}]{chiang2023large}
Cheng-Han Chiang and Hung yi~Lee. 2023.
\newblock \href {https://arxiv.org/abs/2305.01937} {Can large language models be an alternative to human evaluations?}
\newblock \emph{Preprint}, arXiv:2305.01937.

\bibitem[{Cornia et~al.(2020)Cornia, Stefanini, Baraldi, and Cucchiara}]{cornia2020meshed}
Marcella Cornia, Matteo Stefanini, Lorenzo Baraldi, and Rita Cucchiara. 2020.
\newblock Meshed-memory transformer for image captioning.
\newblock In \emph{Proceedings of the IEEE/CVF conference on computer vision and pattern recognition}, pages 10578--10587.

\bibitem[{Dosovitskiy et~al.(2021)Dosovitskiy, Beyer, Kolesnikov, Weissenborn, Zhai, Unterthiner, Dehghani, Minderer, Heigold, Gelly, Uszkoreit, and Houlsby}]{dosovitskiy2021image}
Alexey Dosovitskiy, Lucas Beyer, Alexander Kolesnikov, Dirk Weissenborn, Xiaohua Zhai, Thomas Unterthiner, Mostafa Dehghani, Matthias Minderer, Georg Heigold, Sylvain Gelly, Jakob Uszkoreit, and Neil Houlsby. 2021.
\newblock \href {https://arxiv.org/abs/2010.11929} {An image is worth 16x16 words: Transformers for image recognition at scale}.
\newblock \emph{Preprint}, arXiv:2010.11929.

\bibitem[{Dosovitskiy et~al.(2020)Dosovitskiy, Beyer, Kolesnikov, Weissenborn, Zhai, Unterthiner, Dehghani, Minderer, Heigold, Gelly et~al.}]{dosovitskiy2020image}
Alexey Dosovitskiy, Lucas Beyer, Alexander Kolesnikov, Dirk Weissenborn, Xiaohua Zhai, Thomas Unterthiner, Mostafa Dehghani, Matthias Minderer, Georg Heigold, Sylvain Gelly, et~al. 2020.
\newblock An image is worth 16x16 words: Transformers for image recognition at scale.
\newblock In \emph{International Conference on Learning Representations}.

\bibitem[{Fang et~al.(2022)Fang, Wang, Hu, Liang, Gan, Wang, Yang, and Liu}]{fang2022injecting}
Zhiyuan Fang, Jianfeng Wang, Xiaowei Hu, Lin Liang, Zhe Gan, Lijuan Wang, Yezhou Yang, and Zicheng Liu. 2022.
\newblock Injecting semantic concepts into end-to-end image captioning.
\newblock In \emph{Proceedings of the IEEE/CVF conference on computer vision and pattern recognition}, pages 18009--18019.

\bibitem[{Fu et~al.(2023)Fu, Ng, Jiang, and Liu}]{fu2023gptscore}
Jinlan Fu, See-Kiong Ng, Zhengbao Jiang, and Pengfei Liu. 2023.
\newblock \href {https://arxiv.org/abs/2302.04166} {Gptscore: Evaluate as you desire}.
\newblock \emph{Preprint}, arXiv:2302.04166.

\bibitem[{Hessel et~al.(2021)Hessel, Holtzman, Forbes, Le~Bras, and Choi}]{hessel2021clipscore}
Jack Hessel, Ari Holtzman, Maxwell Forbes, Ronan Le~Bras, and Yejin Choi. 2021.
\newblock Clipscore: A reference-free evaluation metric for image captioning.
\newblock In \emph{Proceedings of the 2021 Conference on Empirical Methods in Natural Language Processing}, pages 7514--7528.

\bibitem[{Holtzman et~al.(2019)Holtzman, Buys, Du, Forbes, and Choi}]{holtzman2019curious}
Ari Holtzman, Jan Buys, Li~Du, Maxwell Forbes, and Yejin Choi. 2019.
\newblock The curious case of neural text degeneration.
\newblock In \emph{International Conference on Learning Representations}.

\bibitem[{Hosang et~al.(2017)Hosang, Benenson, and Schiele}]{hosang2017learning}
Jan Hosang, Rodrigo Benenson, and Bernt Schiele. 2017.
\newblock Learning non-maximum suppression.
\newblock In \emph{Proceedings of the IEEE conference on computer vision and pattern recognition}, pages 4507--4515.

\bibitem[{Ji et~al.(2021)Ji, Chen, and Wang}]{ji2021step}
Zhong Ji, Kexin Chen, and Haoran Wang. 2021.
\newblock Step-wise hierarchical alignment network for image-text matching.
\newblock In \emph{IJCAI}.

\bibitem[{Karpathy and Fei-Fei(2015)}]{karpathy2015deep}
Andrej Karpathy and Li~Fei-Fei. 2015.
\newblock \href {https://arxiv.org/abs/1412.2306} {Deep visual-semantic alignments for generating image descriptions}.
\newblock \emph{Preprint}, arXiv:1412.2306.

\bibitem[{Kirillov et~al.(2023)Kirillov, Mintun, Ravi, Mao, Rolland, Gustafson, Xiao, Whitehead, Berg, Lo, Doll{\'a}r, and Girshick}]{kirillov2023segany}
Alexander Kirillov, Eric Mintun, Nikhila Ravi, Hanzi Mao, Chloe Rolland, Laura Gustafson, Tete Xiao, Spencer Whitehead, Alexander~C. Berg, Wan-Yen Lo, Piotr Doll{\'a}r, and Ross Girshick. 2023.
\newblock Segment anything.
\newblock \emph{arXiv:2304.02643}.

\bibitem[{Li et~al.(2023)Li, Li, Savarese, and Hoi}]{li2023blip}
Junnan Li, Dongxu Li, Silvio Savarese, and Steven Hoi. 2023.
\newblock Blip-2: Bootstrapping language-image pre-training with frozen image encoders and large language models.
\newblock \emph{arXiv preprint arXiv:2301.12597}.

\bibitem[{Li et~al.(2022{\natexlab{a}})Li, Li, Xiong, and Hoi}]{li2022blip}
Junnan Li, Dongxu Li, Caiming Xiong, and Steven Hoi. 2022{\natexlab{a}}.
\newblock Blip: Bootstrapping language-image pre-training for unified vision-language understanding and generation.
\newblock In \emph{International Conference on Machine Learning}, pages 12888--12900. PMLR.

\bibitem[{Li et~al.(2022{\natexlab{b}})Li, Li, Xiong, and Hoi}]{li2022blipbootstrappinglanguageimagepretraining}
Junnan Li, Dongxu Li, Caiming Xiong, and Steven Hoi. 2022{\natexlab{b}}.
\newblock \href {https://arxiv.org/abs/2201.12086} {Blip: Bootstrapping language-image pre-training for unified vision-language understanding and generation}.
\newblock \emph{Preprint}, arXiv:2201.12086.

\bibitem[{Lin(2004)}]{lin-2004-rouge}
Chin-Yew Lin. 2004.
\newblock \href {https://aclanthology.org/W04-1013} {{ROUGE}: A package for automatic evaluation of summaries}.
\newblock In \emph{Text Summarization Branches Out}, pages 74--81, Barcelona, Spain. Association for Computational Linguistics.

\bibitem[{Lin et~al.(2014)Lin, Maire, Belongie, Hays, Perona, Ramanan, Doll{\'a}r, and Zitnick}]{lin2014microsoft}
Tsung-Yi Lin, Michael Maire, Serge Belongie, James Hays, Pietro Perona, Deva Ramanan, Piotr Doll{\'a}r, and C~Lawrence Zitnick. 2014.
\newblock Microsoft coco: Common objects in context.
\newblock In \emph{Computer Vision--ECCV 2014: 13th European Conference, Zurich, Switzerland, September 6-12, 2014, Proceedings, Part V 13}, pages 740--755. Springer.

\bibitem[{Liu et~al.(2023{\natexlab{a}})Liu, Cheng, Liu, Zhang, Li, Ren, Zou, Yang, Su, Zhu et~al.}]{liu2023llava}
Shilong Liu, Hao Cheng, Haotian Liu, Hao Zhang, Feng Li, Tianhe Ren, Xueyan Zou, Jianwei Yang, Hang Su, Jun Zhu, et~al. 2023{\natexlab{a}}.
\newblock Llava-plus: Learning to use tools for creating multimodal agents.
\newblock \emph{arXiv preprint arXiv:2311.05437}.

\bibitem[{Liu et~al.(2023{\natexlab{b}})Liu, Iter, Xu, Wang, Xu, and Zhu}]{liu2023geval}
Yang Liu, Dan Iter, Yichong Xu, Shuohang Wang, Ruochen Xu, and Chenguang Zhu. 2023{\natexlab{b}}.
\newblock \href {https://arxiv.org/abs/2303.16634} {G-eval: Nlg evaluation using gpt-4 with better human alignment}.
\newblock \emph{Preprint}, arXiv:2303.16634.

\bibitem[{Oquab et~al.(2023)Oquab, Darcet, Moutakanni, Vo, Szafraniec, Khalidov, Fernandez, Haziza, Massa, El-Nouby et~al.}]{oquab2023dinov2}
Maxime Oquab, Timoth{\'e}e Darcet, Th{\'e}o Moutakanni, Huy Vo, Marc Szafraniec, Vasil Khalidov, Pierre Fernandez, Daniel Haziza, Francisco Massa, Alaaeldin El-Nouby, et~al. 2023.
\newblock Dinov2: Learning robust visual features without supervision.
\newblock \emph{arXiv preprint arXiv:2304.07193}.

\bibitem[{Papineni et~al.(2002{\natexlab{a}})Papineni, Roukos, Ward, and Zhu}]{bleu}
Kishore Papineni, Salim Roukos, Todd Ward, and Wei-Jing Zhu. 2002{\natexlab{a}}.
\newblock \href {https://doi.org/10.3115/1073083.1073135} {Bleu: A method for automatic evaluation of machine translation}.
\newblock In \emph{Proceedings of the 40th Annual Meeting on Association for Computational Linguistics}, ACL '02, page 311–318, USA. Association for Computational Linguistics.

\bibitem[{Papineni et~al.(2002{\natexlab{b}})Papineni, Roukos, Ward, and Zhu}]{papineni2002bleu}
Kishore Papineni, Salim Roukos, Todd Ward, and Wei-Jing Zhu. 2002{\natexlab{b}}.
\newblock Bleu: a method for automatic evaluation of machine translation.
\newblock In \emph{Proceedings of the 40th annual meeting of the Association for Computational Linguistics}, pages 311--318.

\bibitem[{Radford et~al.(2021{\natexlab{a}})Radford, Kim, Hallacy, Ramesh, Goh, Agarwal, Sastry, Askell, Mishkin, Clark, Krueger, and Sutskever}]{radford2021learningtransferablevisualmodels}
Alec Radford, Jong~Wook Kim, Chris Hallacy, Aditya Ramesh, Gabriel Goh, Sandhini Agarwal, Girish Sastry, Amanda Askell, Pamela Mishkin, Jack Clark, Gretchen Krueger, and Ilya Sutskever. 2021{\natexlab{a}}.
\newblock \href {https://arxiv.org/abs/2103.00020} {Learning transferable visual models from natural language supervision}.
\newblock \emph{Preprint}, arXiv:2103.00020.

\bibitem[{Radford et~al.(2021{\natexlab{b}})Radford, Kim, Hallacy, Ramesh, Goh, Agarwal, Sastry, Askell, Mishkin, Clark et~al.}]{radford2021learning}
Alec Radford, Jong~Wook Kim, Chris Hallacy, Aditya Ramesh, Gabriel Goh, Sandhini Agarwal, Girish Sastry, Amanda Askell, Pamela Mishkin, Jack Clark, et~al. 2021{\natexlab{b}}.
\newblock Learning transferable visual models from natural language supervision.
\newblock In \emph{International conference on machine learning}, pages 8748--8763. PMLR.

\bibitem[{Reimers and Gurevych(2019)}]{reimers2019sentence}
Nils Reimers and Iryna Gurevych. 2019.
\newblock Sentence-bert: Sentence embeddings using siamese bert-networks.
\newblock In \emph{Proceedings of the 2019 Conference on Empirical Methods in Natural Language Processing and the 9th International Joint Conference on Natural Language Processing (EMNLP-IJCNLP)}, pages 3982--3992.

\bibitem[{Selvaraju et~al.(2017)Selvaraju, Cogswell, Das, Vedantam, Parikh, and Batra}]{selvaraju2017grad}
Ramprasaath~R Selvaraju, Michael Cogswell, Abhishek Das, Ramakrishna Vedantam, Devi Parikh, and Dhruv Batra. 2017.
\newblock Grad-cam: Visual explanations from deep networks via gradient-based localization.
\newblock In \emph{Proceedings of the IEEE international conference on computer vision}, pages 618--626.

\bibitem[{Shao et~al.(2023)Shao, Liu, Pei, Xu, Dai, Lu, Li, and Yan}]{Shao2023ICCV}
Bin Shao, Jianzhuang Liu, Renjing Pei, Songcen Xu, Peng Dai, Juwei Lu, Weimian Li, and Youliang Yan. 2023.
\newblock Hivlp: Hierarchical interactive video-language pre-training.
\newblock In \emph{Proceedings of the IEEE/CVF International Conference on Computer Vision (ICCV)}, pages 13756--13766.

\bibitem[{Shlapentokh-Rothman et~al.(2024)Shlapentokh-Rothman, Blume, Xiao, Wu, V, Tao, Lee, Torres, Wang, and Hoiem}]{shlapentokhrothman2024regionbasedrepresentationsrevisited}
Michal Shlapentokh-Rothman, Ansel Blume, Yao Xiao, Yuqun Wu, Sethuraman~T V, Heyi Tao, Jae~Yong Lee, Wilfredo Torres, Yu-Xiong Wang, and Derek Hoiem. 2024.
\newblock \href {https://arxiv.org/abs/2402.02352} {Region-based representations revisited}.
\newblock \emph{Preprint}, arXiv:2402.02352.

\bibitem[{Shukor et~al.(2022)Shukor, Couairon, and Cord}]{shukor2022efficient}
Mustafa Shukor, Guillaume Couairon, and Matthieu Cord. 2022.
\newblock Efficient vision-language pretraining with visual concepts and hierarchical alignment.
\newblock In \emph{33rd British Machine Vision Conference (BMVC)}.

\bibitem[{Suo et~al.(2023)Suo, Zhu, and Yang}]{suo-etal-2023-text}
Yucheng Suo, Linchao Zhu, and Yi~Yang. 2023.
\newblock \href {https://doi.org/10.18653/v1/2023.findings-emnlp.73} {Text augmented spatial aware zero-shot referring image segmentation}.
\newblock In \emph{Findings of the Association for Computational Linguistics: EMNLP 2023}, pages 1032--1043, Singapore. Association for Computational Linguistics.

\bibitem[{Tiong et~al.(2022)Tiong, Li, Li, Savarese, and Hoi}]{tiong2022plug}
Anthony Meng~Huat Tiong, Junnan Li, Boyang Li, Silvio Savarese, and Steven~CH Hoi. 2022.
\newblock Plug-and-play vqa: Zero-shot vqa by conjoining large pretrained models with zero training.
\newblock In \emph{Findings of the Association for Computational Linguistics: EMNLP 2022}, pages 951--967.

\bibitem[{Touvron et~al.(2023)Touvron, Martin, Stone, Albert, Almahairi, Babaei, Bashlykov, Batra, Bhargava, Bhosale et~al.}]{touvron2023llama}
Hugo Touvron, Louis Martin, Kevin Stone, Peter Albert, Amjad Almahairi, Yasmine Babaei, Nikolay Bashlykov, Soumya Batra, Prajjwal Bhargava, Shruti Bhosale, et~al. 2023.
\newblock Llama 2: Open foundation and fine-tuned chat models.
\newblock \emph{arXiv preprint arXiv:2307.09288}.

\bibitem[{Vedantam et~al.(2015)Vedantam, Zitnick, and Parikh}]{cider}
Ramakrishna Vedantam, C.~Lawrence Zitnick, and Devi Parikh. 2015.
\newblock \href {https://doi.org/10.1109/CVPR.2015.7299087} {Cider: Consensus-based image description evaluation}.
\newblock In \emph{2015 IEEE Conference on Computer Vision and Pattern Recognition (CVPR)}, pages 4566--4575.

\bibitem[{Waheed et~al.(2025)Waheed, An, Milford, Ramchurn, and Ehsan}]{waheed2025vlm}
Sania Waheed, Na~Min An, Michael Milford, Sarvapali~D Ramchurn, and Shoaib Ehsan. 2025.
\newblock Vlm-guided visual place recognition for planet-scale geo-localization.
\newblock \emph{arXiv preprint arXiv:2507.17455}.

\bibitem[{Wang et~al.(2022{\natexlab{a}})Wang, Bao, Dong, Bjorck, Peng, Liu, Aggarwal, Mohammed, Singhal, Som, and Wei}]{wang2022imageforeignlanguagebeit}
Wenhui Wang, Hangbo Bao, Li~Dong, Johan Bjorck, Zhiliang Peng, Qiang Liu, Kriti Aggarwal, Owais~Khan Mohammed, Saksham Singhal, Subhojit Som, and Furu Wei. 2022{\natexlab{a}}.
\newblock \href {https://arxiv.org/abs/2208.10442} {Image as a foreign language: Beit pretraining for all vision and vision-language tasks}.
\newblock \emph{Preprint}, arXiv:2208.10442.

\bibitem[{Wang et~al.(2022{\natexlab{b}})Wang, Bao, Dong, Bjorck, Peng, Liu, Aggarwal, Mohammed, Singhal, Som et~al.}]{wang2022image}
Wenhui Wang, Hangbo Bao, Li~Dong, Johan Bjorck, Zhiliang Peng, Qiang Liu, Kriti Aggarwal, Owais~Khan Mohammed, Saksham Singhal, Subhojit Som, et~al. 2022{\natexlab{b}}.
\newblock Image as a foreign language: Beit pretraining for all vision and vision-language tasks.
\newblock \emph{arXiv preprint arXiv:2208.10442}.

\bibitem[{Wang et~al.(2025)Wang, Ni, Liu, Yuan, and Tang}]{wang2025iterprime}
Yuji Wang, Jingchen Ni, Yong Liu, Chun Yuan, and Yansong Tang. 2025.
\newblock Iterprime: Zero-shot referring image segmentation with iterative grad-cam refinement and primary word emphasis.
\newblock \emph{arXiv preprint arXiv:2503.00936}.

\bibitem[{Xu et~al.(2021)Xu, Niu, Tan, Luo, Du, and Wu}]{xu2021towards}
Guanghui Xu, Shuaicheng Niu, Mingkui Tan, Yucheng Luo, Qing Du, and Qi~Wu. 2021.
\newblock Towards accurate text-based image captioning with content diversity exploration.
\newblock In \emph{Proceedings of the IEEE/CVF Conference on Computer Vision and Pattern Recognition}, pages 12637--12646.

\bibitem[{Yao et~al.(2019)Yao, Pan, Li, and Mei}]{yao2019hierarchy}
Ting Yao, Yingwei Pan, Yehao Li, and Tao Mei. 2019.
\newblock Hierarchy parsing for image captioning.
\newblock In \emph{Proceedings of the IEEE/CVF international conference on computer vision}, pages 2621--2629.

\bibitem[{Young et~al.(2014)Young, Lai, Hodosh, and Hockenmaier}]{young2014image}
Peter Young, Alice Lai, Micah Hodosh, and Julia Hockenmaier. 2014.
\newblock From image descriptions to visual denotations: New similarity metrics for semantic inference over event descriptions.
\newblock \emph{Transactions of the Association for Computational Linguistics}, 2:67--78.

\bibitem[{Yu et~al.(2023)Yu, Seo, and Son}]{Yu_2023_CVPR}
Seonghoon Yu, Paul~Hongsuck Seo, and Jeany Son. 2023.
\newblock Zero-shot referring image segmentation with global-local context features.
\newblock In \emph{Proceedings of the IEEE/CVF Conference on Computer Vision and Pattern Recognition (CVPR)}, pages 19456--19465.

\bibitem[{Yuksekgonul et~al.(2022)Yuksekgonul, Bianchi, Kalluri, Jurafsky, and Zou}]{yuksekgonul2022and}
Mert Yuksekgonul, Federico Bianchi, Pratyusha Kalluri, Dan Jurafsky, and James Zou. 2022.
\newblock When and why vision-language models behave like bags-of-words, and what to do about it?
\newblock In \emph{The Eleventh International Conference on Learning Representations}.

\bibitem[{Zhang et~al.(2021)Zhang, Sun, Luo, Ji, Zhou, Wu, Huang, and Ji}]{zhang2021rstnet}
Xuying Zhang, Xiaoshuai Sun, Yunpeng Luo, Jiayi Ji, Yiyi Zhou, Yongjian Wu, Feiyue Huang, and Rongrong Ji. 2021.
\newblock Rstnet: Captioning with adaptive attention on visual and non-visual words.
\newblock In \emph{Proceedings of the IEEE/CVF conference on computer vision and pattern recognition}, pages 15465--15474.

\bibitem[{Zhao et~al.(2023)Zhao, Ding, An, Du, Yu, Li, Tang, and Wang}]{zhao2023fast}
Xu~Zhao, Wenchao Ding, Yongqi An, Yinglong Du, Tao Yu, Min Li, Ming Tang, and Jinqiao Wang. 2023.
\newblock \href {https://arxiv.org/abs/2306.12156} {Fast segment anything}.
\newblock \emph{Preprint}, arXiv:2306.12156.

\end{thebibliography}

\appendix
\onecolumn
\section{Appendix}

\section{Additional Related Works}

\subsection{Vision-Language Models (VLMs)}
A growth of interest in VLMs has continued due to the wide availability of multimodal data on the web \cite{waheed2025vlm}. Foundation VLMs can be applied to a range of tasks in a zero-shot manner. Notably, CLIP \cite{radford2021learning} jointly pre-trains an image encoder and a text encoder by maximizing and minimizing the cosine similarity of correct and incorrect image-text pair embeddings, respectively, with image-text contrastive (ITC) loss. In contrast, BLIP \cite{li2022blip} uses both ITC and image-text matching (ITM) loss for enhanced image-text data representation. Additionally, the BLIP \cite{li2022blip} \textit{captioner} uses language modeling (LM) loss for autoregressive image caption generation along with a filter, \textit{capfilt}, to improve the quality of image-text pairs for training.  

Flamingo \cite{alayrac2022flamingo} shows remarkable zero-shot ability in image captioning, visual question-answering (VQA), and image-text retrieval (ITR) tasks by leveraging the few-shot learning ability of pre-trained vision-only and language-only models. It simply interleaves input visual data with task-specific text examples, producing free-form texts for unseen visual data. Another general-purpose model, BEIT3 \cite{wang2022image} with a Multiway Transformer structure, uses different types of modality experts to perform fusion and modality-specific training. A masked modeling objective on images only and image-text pairs is performed for computer vision tasks (\textit{e.g.}, image classification, semantic segmentation, object detection) and vision-language tasks (\textit{e.g.}, VQA), respectively. Whereas the VQA task uses a fused encoder for image-text pairs, the ITR task encodes images and texts independently with ITC loss. Lastly, sequence-to-sequence learning is applied to generate texts from images for the image captioning task. Inspired by these previous works, we propose a meta-VLM model that utilizes a pre-trained BLIP \cite{li2022blip} image captioning module to generate enhanced textual representations, which can later serve as useful data for various downstream tasks.

\subsection{Hierarchical Representation}\label{rw:hier}
Identifying and extracting regions of interest within images is crucial for a hierarchical representation. The most intuitive way to achieve this would typically 
involve the use of object detectors \cite{yao2019hierarchy,cornia2020meshed,zhang2021rstnet}. However, the heavy computational demands of the object detectors inevitably lead to inefficiency during the inference stage \cite{yao2019hierarchy,cornia2020meshed,zhang2021rstnet}. In response, recent works sought to replace these cumbersome detectors by adopting visual concepts in the form of object tags \cite{fang2022injecting,shukor2022efficient} as an alternative. However, this detector-free approach is contingent upon the availability of object-specific data within the dataset. Employing pre-trained models is a more efficient way to identify areas of interest within images. GradCAM \cite{selvaraju2017grad} highlights essential regions that the pre-trained models use to predict any target concept using its gradients with respect to feature map activations of the final convolutional layer. DINOv2 \cite{oquab2023dinov2} capitalizes on existing self-supervised pre-trained models to generate robust, all-purpose visual features, supporting a wide array of tasks ranging from image-level classification to pixel-level segmentation. However, the image regions/features delineated by GradCAM/DINOv2 tend to show saliency for specific tasks and are unable to capture the full spectrum of visual representations. Conversely, SAM \cite{kirillov2023segany} intricately segments every semantically significant component of an image into high-quality segmentation masks generated by prompting with various inputs such as point, box, mask, or free-form text, unrestricted by the types of tasks. In our framework, we integrate SAM \cite{kirillov2023segany} to create semantically meaningful segmentation masks for an entire image automatically.

Several prior studies have incorporated the principles of hierarchy or multi-scale representation into their model architectures, aiming to enhance the alignment between images and texts \cite{ji2021step,Shao2023ICCV,shukor2022efficient}. SHAN \cite{ji2021step} deconstructs the image-text matching process into two distinct facets: fragment-level and context-level alignments, enabling matches across three different scopes: local-to-local, global-to-local, and global-to-global. HiVLP \cite{Shao2023ICCV} leverages both low- and high-dimensional features to represent coarse and fine details. ViCHA \cite{shukor2022efficient} aligns images and texts across various layers of neural network encoders with the underlying assumption that each layer reflects varying semantic levels. Unlike these approaches, we divide the segmentation masks hierarchically and use the embeddings of the extracted individual image patches for caption generation.

\subsection{Caption Evaluation}\label{rw:eval}
Common image captioning evaluation metrics, including BLEU \cite{bleu}, METEOR \cite{banerjee-lavie-2005-meteor}, ROUGE \cite{lin-2004-rouge}, and CIDEr \cite{cider} scores, are primarily n-gram approaches that assess the quality of generated captions by considering their overlap with human-generated captions. Most SOTA VLMs frequently exhibit promising scores across these conventional evaluation metrics. However, these metrics are limited in their capabilities to measure the diversity of the generated captions. This limitation leads to a bias in these models towards generating an "average" and "safe" caption reflecting the most basic information in the image, rendering them less informative than human-generated captions. To address this gap, we incorporate several diversity metrics, including mBLEU-4, Div-2 \cite{aneja2019sequential}, and the proposed pairwise cosine distance (PCD), along with semantic integrity and relevance scores to ensure that the captions generated by our framework are not only diverse but also meaningful and directly relevant to the given image and human-annotated captions.

\section{Experiments}
\label{app:exp}
 
\subsection{Implementation Details} \label{app:impdetails}
 The ISM (Section~\ref{method:ims}) employs the fully automated SAM with no prompting \cite{kirillov2023segany}, along with the image encoder initialized from ViT (\verb|ViT-L/16|) pre-trained on ImageNet \cite{dosovitskiy2021image}, following the same settings as BLIP \cite{li2022blip}. Note that we use BLIP \cite{li2022blip} for captioning instead of BLIP-2 \cite{li2023blip} since BLIP-2 uses intermediate representations trained on pairs of entire images and texts for caption generation using an LLM, which is not directly applicable to HBoP that uses pairs of image patches and texts.

 The HCM (Section~\ref{method:hcm}) creates the global level by selecting the top ($k=5$) masks with the largest areas and designating the remaining masks as fine-grained. To create the regional level, $K$-means clustering, with ($K=5$) clusters per image, is applied to the bounding boxes of the segmentation masks. For each cluster, we extract patch embeddings corresponding to the semantically segmented regions, projecting the clusters onto the image to identify relevant patches. The embeddings from each cluster are concatenated, and zero-padding is applied to preserve the original shape, ensuring that each cluster represents a semantically coherent region of the image. These regional embeddings are then used as input for caption generation in the captioning module. Lastly, the ICM (Section~\ref{method:icm}) follows the methodology outlined in \citealp{tiong2022plug}.
 
 While NMS with a threshold of 0.1 is applied at all three levels for all main experiments, we also conduct ablation experiments to examine the impact of varying the IoU threshold used in our sampling strategy. As shown in Table~\ref{tab:iou_results},  while performance shows slight variation across different thresholds, our method consistently outperforms the baselines. This variance occurs because increasing the IoU threshold allows more overlapping image regions to be included, which tends to generate more repetitive captions and thus reduces diversity (PCD and Div-2 scores decreased).

 \begin{table}[h!]
\centering
\resizebox{0.35\textwidth}{!}{
\begin{tabular}{c|c|c}
\toprule
IoU Threshold & PCD & Div-2 ↑ \\
\midrule
0.1 & 0.772 & 0.735 \\
0.4 & 0.709 & 0.728 \\
0.8 & 0.684 & 0.666 \\
\bottomrule
\end{tabular}
}
\caption{Diversity results at different IoU thresholds.}
\label{tab:iou_results}
\end{table}
 
 Although HBoP presents a three-tier hierarchical structure, it is crucial to note that we adjust the different hierarchy levels depending on a given dataset. A dataset with information-rich complex images would require using all three hierarchy levels. However, a dataset with relatively simpler images, such as the MSCOCO dataset \cite{lin2014microsoft}, would benefit from a two-tier hierarchy with just the global and regional captions. We use the first two levels during evaluations unless specified otherwise.

 All the model captions in Tables~\ref{table:Div} and \ref{table:Div_nocaps} are regenerated, except for Seq-CVAE \cite{aneja2019sequential}, where the results are taken directly from the original paper. While HBoP benefits from bounding box information, it is important to note that other baseline methods (\textit{e.g.}, ModeCap) have the additional advantage of explicit learning objectives to improve diversity. The exact prompts we use for Honeybee \cite{cha2023honeybee} (top) and LLaVA-1.5/1.6 \cite{liu2023llava} are in Table~\ref{app:prompt_gen}.

\subsection{Crop vs. Embedding Sampling Comparison}
We present a comparison between our embedding-sampling approach (HBoP) and a baseline where segmented image regions are directly cropped and captioned using BLIP. Table~\ref{tab:crop} shows that while cropping can improve diversity scores, it often sacrifices relevance as indicated by SBERT. In contrast, HBoP preserves contextual understanding by sampling from full-image embeddings.

\begin{table}[h]
\centering
\small
\begin{tabular}{lccc|ccc}
\toprule
\multirow{2}{*}{Models} & \multicolumn{3}{c|}{\textbf{MSCOCO}} & \multicolumn{3}{c}{\textbf{Flickr30k}} \\
              & SBERT ↑ & mBLEU-4 ↓ & Div-2 ↑ & SBERT ↑ & mBLEU-4 ↓ & Div-2 ↑ \\
\midrule
BLIP (-NS) & 56.00 & 1.00 & 0.179 & 55.78 & 1.00 & 0.179\\
BLIP (+NS) & 57.23 & 0.66 & 0.387 & 46.99 & 0.66 & 0.387\\
Crop       & 52.03 & 0.10 & 0.600 & 50.00 & 0.08 & 0.610\\
\textbf{HBoP} & 56.30 & 0.05 & 0.735 & 54.00 & 0.04 & 0.750\\
\bottomrule
\end{tabular}
\caption{Comparison of our embedding sampling approach (HBoP) with direct cropping of segmented regions.}
\label{tab:crop}
\end{table}

\subsection{Evaluation}
We evaluate the model captions using three distinct metrics: 1) diversity across captions per image, 2) relevancy with images, and 3) semantic coherence and meaningfulness. The datasets we use for evaluation are: the Karpathy test split \cite{karpathy2015deep} of MSCOCO (5k images) \cite{lin2014microsoft}, Flickr30K zero-shot (1k test images) \cite{young2014image}, and NoCaps validation (4.5k images) \cite{agrawal2019nocaps}.

\subsubsection{Diversity}
 We measure the diversity in the generated captions using the cosine similarity between the sentence embeddings of all the corresponding captions per image. The comparison baselines are random captions, where each caption corresponds to different images, BLIP \cite{li2022blip} with and without nucleus sampling (NS\footnote{Unless otherwise specified, all the BLIP models in this paper refer to BLIP with NS.}) \cite{holtzman2019curious}, BLIP-2 \cite{li2023blip}, ModeCap \cite{chen2022learning}, Honeybee \cite{cha2023honeybee}, and gold captions\footnote{We exclude PnP-VQA since the captions are generated per question instead of per image, unlike other baselines.}. The diversity of the generated captions ($s_1, s_2, ... s_n$) per dataset instance\footnote{Note that $n = 5$ for all dataset instances, and we use one global caption and five regional captions for HBoP.} is measured using pairwise cosine distance (PCD):
 
\begin{equation}\label{eq:6}
    \text{PCD}(s_1, s_2, ... s_n) = \frac{1}{n}\sum_{i=1}^n \sum_{j=1}^{j<i} (1 - cos(M(s_i), M(s_j)))
\end{equation}

In the above equation, $cos$ represents the cosine similarity of the input embeddings. We use sentence embeddings from a pre-trained sentence transformer model (\verb|all-MiniLM-L6-v2|) \cite{reimers2019sentence}, denoted as $M$ in Eq.~\ref{eq:6}, that can capture the semantic relationships between captions \cite{an-etal-2024-capturing}. This measure evaluates the extent to which the generated captions differ from each other per image. We report the final diversity score for each dataset as the average PCD score of all images in the dataset. Ideally, the PCD score should be lower than that of random captions that serve as the upper bound of the diversity score, but it should be higher than that for captions generated by existing baselines.

Additionally, we use mBLEU-4 and n-gram diversity (\textit{e.g.}, Div-1, Div-2) \cite{aneja2019sequential}, to compare with more challenging baseline models, such as ModeCap \cite{chen2022learning} and Seq-CVAE \cite{aneja2019sequential} that are built to achieve diversity within captions per image. For ModeCap \cite{chen2022learning}, we follow the default settings from the original paper to reproduce the results based on training the Transformer-DML model. We also prompt a recently introduced multimodal LLM called Honeybee \cite{cha2023honeybee} as follows: "Describe this image with 5 diverse captions, using less than 20 words for each caption."

\subsubsection{Relevancy} While confirming that each dataset contains captions with high semantic integrity is crucial, the captions must also be relevant to the corresponding images. We employ CLIP-Score \cite{hessel2021clipscore} that calculates the correlation between visual and textual CLIP embeddings \cite{radford2021learning} using pre-trained ViT (\verb|openai/| \verb|clip-vit-base-patch32|) without relying on human-generated references. Similar to the comparison baseline datasets for semantic integrity evaluation, we compare HBoP with PnP-VQA \cite{tiong2022plug}, BLIP \cite{li2022blip}, BLIP-2 \cite{li2023blip}, gold captions, and random captions. We generate random captions by selecting five random captions for each image from a pool of HBoP captions corresponding to different images. In other words, although the random caption itself should make sense, they depict mismatched images. We randomly select one out of a total of five captions per image for each dataset and compute the correlation between CLIPScores of generated captions and gold captions.

Additionally, we measure the semantic similarity between ground-truth (or \textit{gold}) captions and captions generated with models using transformer-based SBERT \cite{reimers2019sentence}. Note that this metric is robust to synonyms or paraphrasing, unlike n-gram metrics \cite{papineni2002bleu, lin-2004-rouge}.

\subsubsection{Semantic Integrity}\label{app:semantic}
\noindent
\textbf{HBoP generates semantically meaningful captions.} We evaluate the semantic integrity of HBoP captions using LLMs, LLaMA-2-13b \cite{touvron2023llama} and GPT-4 \cite{fu2023gptscore}, which have shown high correlation with human judgment \cite{chiang2023large,liu2023geval,fu2023gptscore}. Table~\ref{tab:2} shows that HBoP achieves semantic integrity scores close to the gold captions and notably outperforms models like PnP-VQA \cite{tiong2022plug}. We attribute this improvement to our method’s ability to sample more meaningful image embeddings via the proposed Hierarchical Composition Module (HCM).

We prompt Llama-2-13B (\verb|Llama-2-13b-chat| \verb|-hf|) \cite{touvron2023llama} to access the semantic integrity of HBoP captions along with gold and other baselines (PnP-VQA \cite{tiong2022plug}, BLIP \cite{li2022blip}, BLIP-2 \cite{li2023blip}) captions. Specifically, we randomly select two captions out of a total of five captions per image for each dataset and evaluate the semantic integrity by averaging the coherency and meaningfulness scores for each caption using the prompt shown in Table~\ref{app:prompt_eval}. We use the prompt "This is a picture of" to generate captions for all models in our experiments. This deliberate choice ensures a fair comparison of the general caption generation ability across models, as altering the prompt can yield significantly different results, making fair evaluation challenging. 

Similarly, we use GPT-4 \cite{fu2023gptscore} for additional Semantic Integrity evaluation using only a single caption per image with the prompt shown in Table~\ref{app:prompt_gpteval}. Note that we sample the first 1k image instances in each dataset for this evaluation due to the cost limitations.

\begin{table}[h!]
\begin{center}
\resizebox{0.5\columnwidth}{!}{%
\begin{tabular}{l|cccc|c}
\toprule
 & PnP-VQA & BLIP & BLIP-2 & Gold & HBoP \\
 \midrule
LLama-2-13B & \begin{tabular}[c]{@{}c@{}}7.70\\($\pm$0.09)\end{tabular} & \begin{tabular}[c]{@{}c@{}}9.36\\ ($\pm$0.05)\end{tabular} & \begin{tabular}[c]{@{}c@{}}9.69\\ ($\pm$0.05)\end{tabular} & \begin{tabular}[c]{@{}c@{}}9.17\\ ($\pm$0.06)\end{tabular} & \begin{tabular}[c]{@{}c@{}}8.56\\ ($\pm$0.07)\end{tabular} \\

GPT-4 & \begin{tabular}[c]{@{}c@{}}2.18\\ ($\pm$0.84)\end{tabular} & \begin{tabular}[c]{@{}c@{}}2.97\\ ($\pm$0.10)\end{tabular} & \begin{tabular}[c]{@{}c@{}}2.96\\ ($\pm$0.19)\end{tabular} & \begin{tabular}[c]{@{}c@{}}2.94\\ ($\pm$0.49)\end{tabular} & \begin{tabular}[c]{@{}c@{}}2.48\\ ($\pm$0.73)\end{tabular}  \\
\bottomrule
\end{tabular}}
\caption{Semantic Integrity scores exhibit a similar trend across two LLM evaluations for the Flickr30K dataset (1k test set).}
\label{tab:2}
\end{center}
\end{table}

\section{Additional Results}

\subsection{Relevancy}
\begin{figure*}[t!]
\centering
\includegraphics[width=\textwidth]{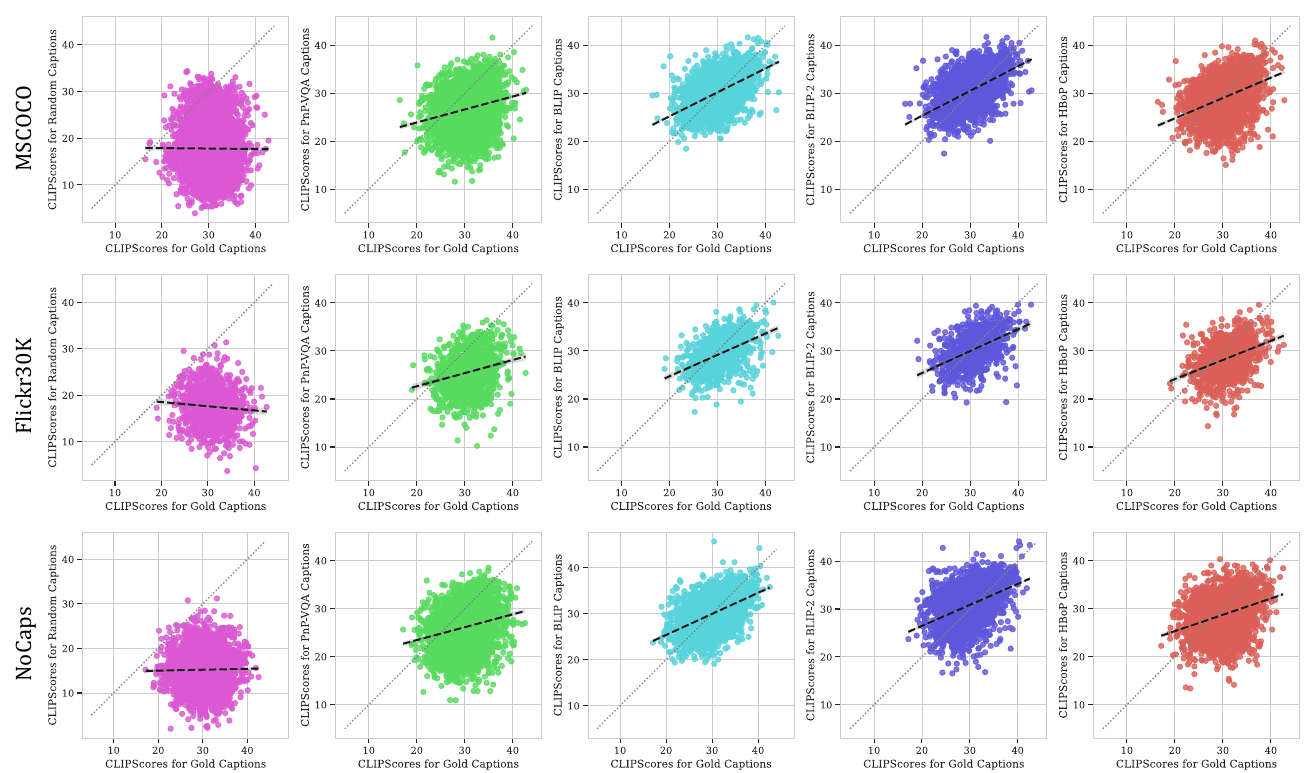}
\caption{Correlation of relevance scores between gold captions and model captions. We observe higher correlations for HBoP, BLIP, and BLIP-2 captions as comapred to random and PnP-VQA captions.}
\label{fig:rel}
\end{figure*}

In Figure~\ref{fig:rel}, HBoP captions (y-axis values in the last column) show comparable relevance scores with gold captions (x-axis values in the last column) with the slope of a linear regression line\footnote{The p-values for all the regression lines are less than 0.001, except for the those of lines in the first columns, which are not statistically significant} being close to 0.5. Although the slopes of these regression lines (MSCOCO \cite{lin2014microsoft}: 0.42, Flickr30k \cite{young2014image}: 0.39, Nocaps \cite{agrawal2019nocaps}: 0.34) are less than those of BLIP \cite{li2022blip} (0.49, 0.44, and 0.45) and BLIP-2 \cite{li2023blip} (0.51, 0.45, 0.43), we observe a trend of having relevance scores in the range of 20 to 40 for both x and y axes values. On the other hand, relevance scores for random and PnP-VQA \cite{tiong2022plug} captions have a spurious and less-correlated relation with those of gold captions.

\subsection{GradCAM Results}
\begin{figure*}[t!]
\centering
\includegraphics[width=\textwidth]{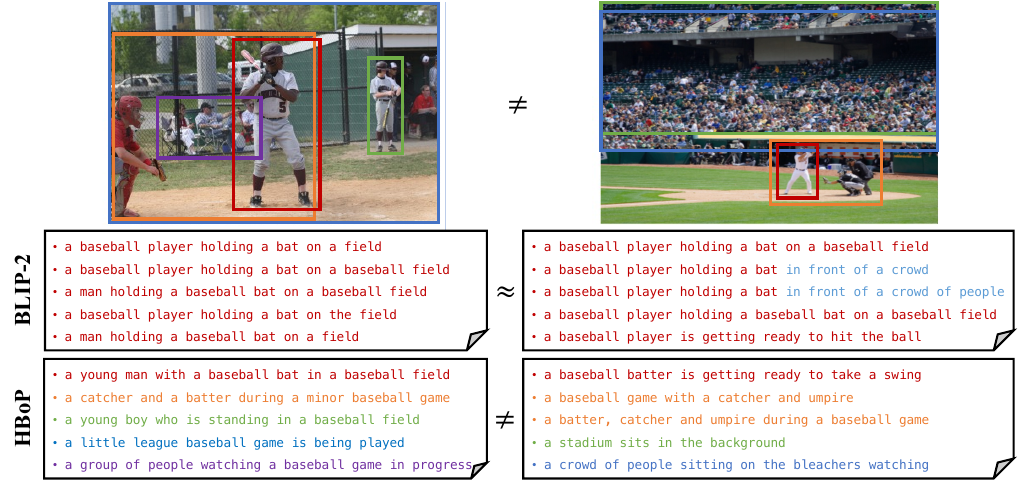}
\caption{Comparison between captions generated using BLIP-2 \cite{li2023blip} and HBoP. Our captions contain more diverse interpretations of the images while maintaining high relevancy.}
\label{fig:motivation}
\end{figure*}

\begin{figure*}[t!]
\centering
\includegraphics[width=\textwidth]{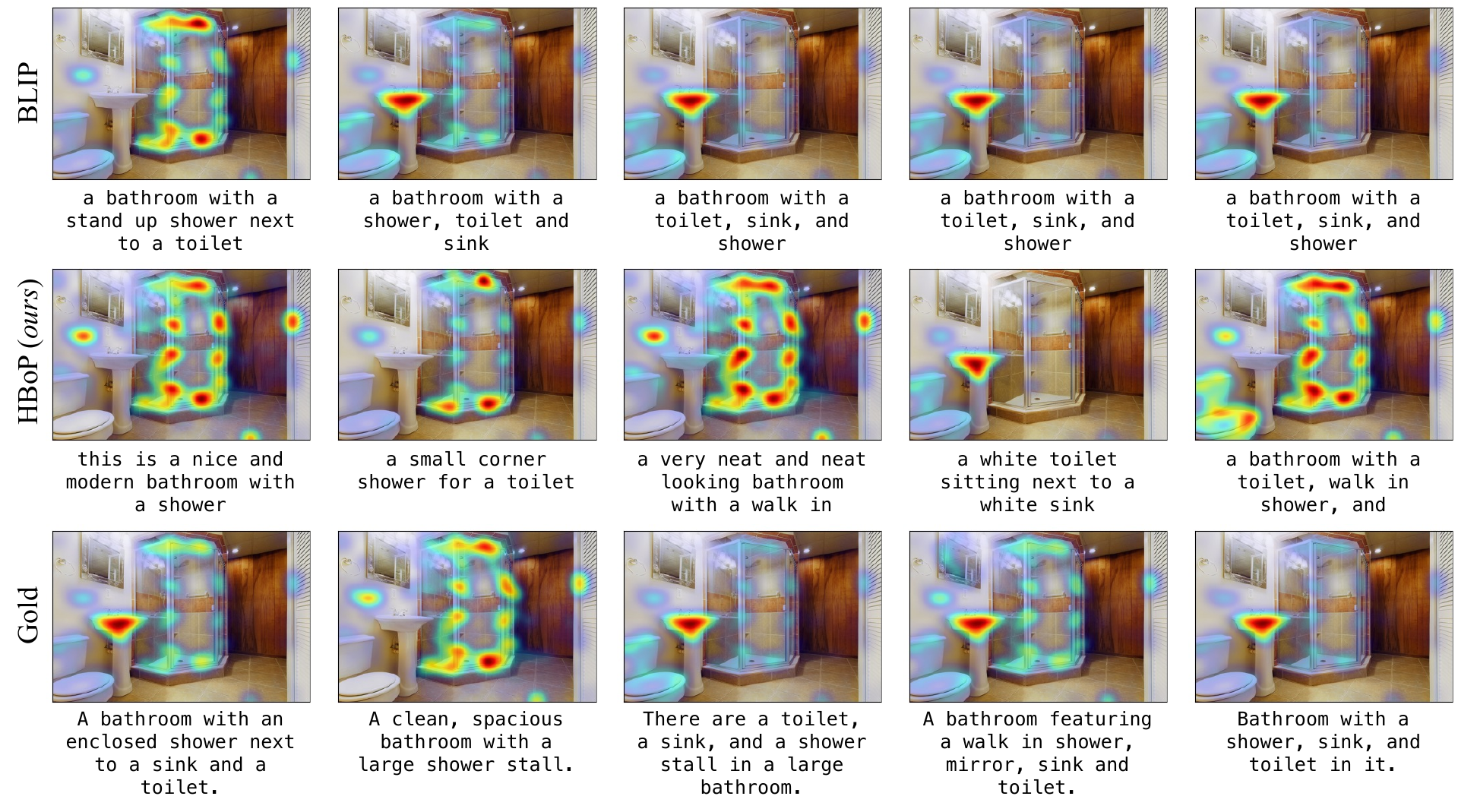}
\includegraphics[width=\textwidth]{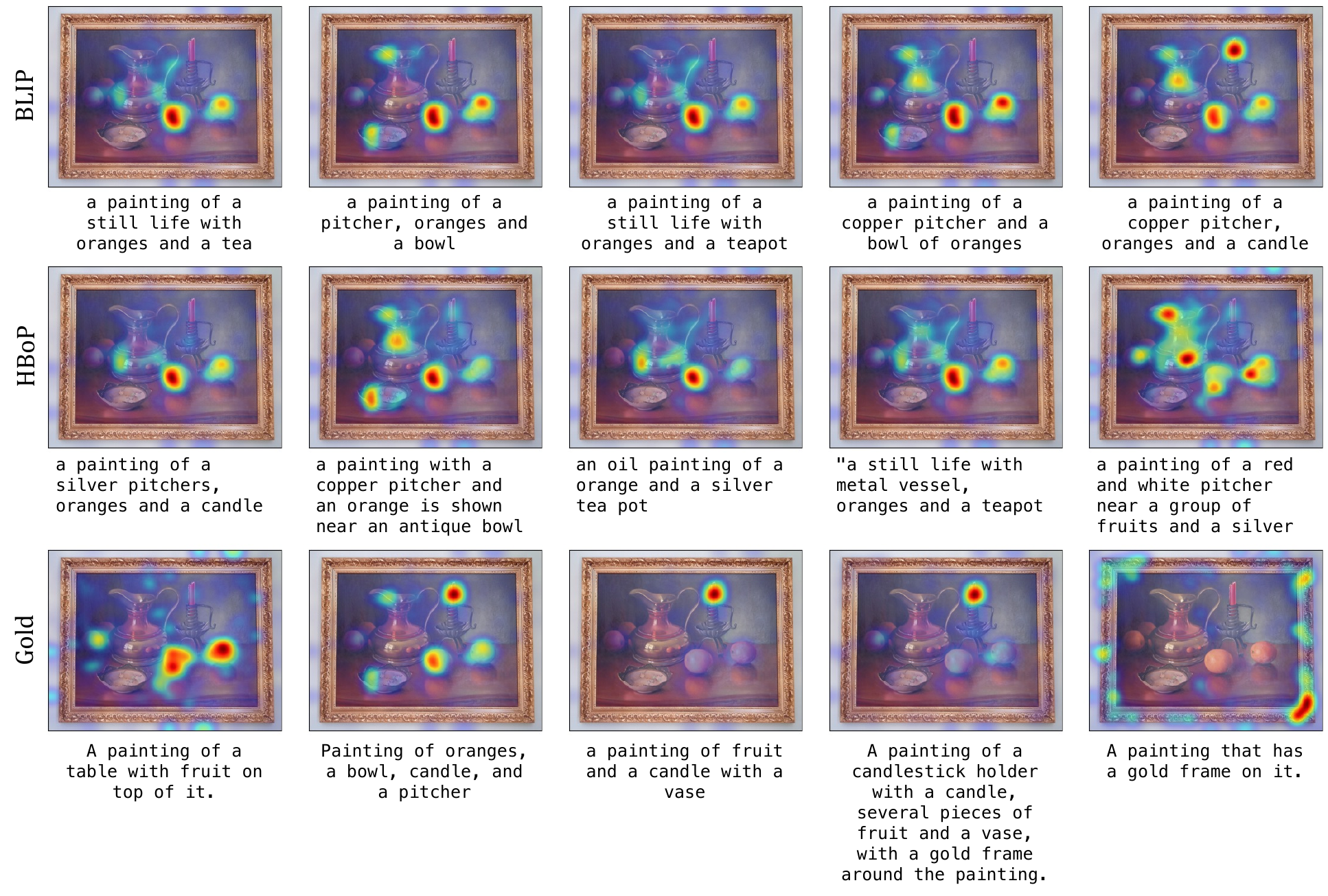}
\caption{Additional visualizations of GradCAMs across different model captions.}
\label{fig:Div}
\end{figure*}

In addition to the evaluation results of the generated captions (samples in Figure~\ref{fig:motivation}), we illustrate how the generated captions correlate with specific image regions through GradCAMs \cite{selvaraju2017grad}. The visual representation identifies the image regions on which the generated captions are based. Specifically, we aggregate the gradients from all cross-attention layers of the pre-trained ITM model in PnP-VQA \cite{tiong2022plug}. Whereas PnP-VQA \cite{tiong2022plug} feeds the question for the textual input, we input BLIP \cite{li2022blip} and gold captions, along with HBoP captions. As shown in Figures \ref{fig:BLIP-2vsHBoP} and \ref{fig:Div}, the highlighted regions in the image for HBoP captions closely resemble the same pattern as those observed using human-generated captions. On the contrary, BLIP exhibits a more constrained range, predominantly concentrating on specific image regions.

\subsection{Fine-grained Captions}\label{app:fine}
Although not evaluated in the perspectives of the three main evaluation metrics, we can also create what we refer to as fine-grained captions that can serve as image tags using our proposed methodology. These serve as supplementary information, enhancing the depth of understanding of the image. They are more vital when dealing with complex images containing various small or intricate objects, which conventional caption generation processes may often overlook. By introducing the additional layer of granularity, our approach ensures a more detailed and inclusive interpretation of the image.

\subsection{FastSAM Results}\label{app:fastsam}
While the segmentation step introduces additional overhead, the hierarchical embedding process itself does not, as it operates directly on the single full-image embedding. Therefore, embedding is not a computational bottleneck in our pipeline. As shown below (Table~\ref{tab:seg_caption_time}), we provide a comparison of inference time (average per image) between BLIP and HBoP on the MSCOCO dataset. While SAM does increase inference time, it brings significant gains in diversity without sacrificing performance. Furthermore, this overhead can be drastically reduced by replacing SAM with FastSAM \cite{zhao2023fast}, which provides comparable segmentation quality at 50–170× faster runtime.

\begin{table}[h!]
\centering
\resizebox{0.8\columnwidth}{!}{%
\begin{tabular}{l|cc}
\toprule
\textbf{Models} & \textbf{Segmentation Time} (s/img) & \textbf{Captioning Time} (s/caption) \\
\midrule
BLIP & -- & 0.38 \\
HBoP (SAM) & 5.43 & 0.38 \\
HBoP (FastSAM) & 0.18 & 0.38 \\
\bottomrule
\end{tabular}
}
\caption{Average segmentation and captioning time for different models.}
\label{tab:seg_caption_time}
\end{table}

\begin{table}[t!]
\begin{center}
\resizebox{0.6\columnwidth}{!}{%
\begin{tabular}{l|c|ccc}
\toprule
 & \# Param & PCD & mBLEU-4 $\downarrow$ & Div-2 $\uparrow$ \\
\midrule
\textit{Random} & - & 0.962 (+0.223) & 0.001 & 0.867 \\
\midrule
BLIP ($-$NS) & 446M & 0.600 (-0.129) & 1.000 & 0.178  \\
BLIP-2 & 3.9B & 0.654 (-0.075) & 0.715 & 0.340  \\
BLIP ($+$NS)  & 446M &  0.679 (-0.050) & 0.629 & 0.400 \\
Honeybee  & 7B & 0.791 (+0.062) & 0.080 & 0.705 \\
\textit{Gold} & - & 0.729 & 0.078 & 0.666 \\
\midrule
HBoP (\textit{ours}) & 1B & 0.783 (+0.054) & 0.041 & 0.748  \\
\multicolumn{2}{c|}{HBoP \textit{Ranking}} & 2/6 & 2/7 & 2/7 \\
\bottomrule
\end{tabular}
}
\caption{Diversity scores for Nocaps test set. We observe a similar diversity trend across model captions as Table~\ref{table:Div}.}
\label{table:Div_nocaps}
\end{center}
\end{table}

\begin{table*}[t!]
\label{tab:prompt}
\begin{tcolorbox}[colback=white!5!white,colframe=black!75!white,arc=0pt,outer arc=0pt]
The following is a conversation between a curious human and AI assistant. The assistant gives helpful, detailed, and polite answers to the user's questions.

Human: <image>

Human: Describe this image with 5 captions with numberings.

AI:

\end{tcolorbox}
\begin{tcolorbox}[colback=white!5!white,colframe=black!75!white,arc=0pt,outer arc=0pt]
A chat between a curious human and an artificial intelligence assistant. The assistant gives helpful, detailed, and polite answers to the human's questions.

Human: <im\_start><image><im\_end>

Human: Describe this image with 5 captions.\#\#\#Assistant:

\tcblower

[INST] <image>
What is shown in this image? Describe this image with 5 captions. [/INST]

\end{tcolorbox}
\caption{Image caption generation prompts for Honeybee (top) and LLaVA-1.5/1.6 (bottom).}
\label{app:prompt_gen}
\end{table*}

\begin{table*}[t!]
\begin{tcolorbox}[colback=white!5!white,colframe=black!75!white,arc=0pt,outer arc=0pt]
[INST] $<<$SYS$>>$ \\
You will be given a caption generated from an image. Given the criteria and rating options, rate the response. Respond with a number only. \\
Evaluation Criteria: \textbf{[CRITERION]}: \textbf{[DEFINITION]} \\
Scale: from 1 to 10 \\
Answer:
$<<$/SYS$>>$ \\
INPUT [$/$INST] \\
\tcblower 
\textbf{[CRITERION]}: Coherence/Meaningfulness \\
\textbf{[DEFINITION]}: the logical and clear connection between ideas or elements within a context. It is characterized by the consistency, integrity, and clarity of information or arguments presented./the relevance and significance of the content in the caption. A meaningful caption goes beyond a literal description, providing insight, context, or emotion that enhances the viewer's understanding or appreciation of the image.
\end{tcolorbox}
\caption{The prompt for evaluating semantic integrity (coherence + meaningfulness) of generated model captions using Llama-2-13B.}
\label{app:prompt_eval}
\end{table*}

\begin{table*}[t!]
\begin{tcolorbox}[colback=white!5!white,colframe=black!75!white,arc=0pt,outer arc=0pt]
You will be given one caption written for describing an image. \\
\\
Your task is to rate the caption on one metric. \\
\\
Please make sure you read and understand these instructions carefully. Please keep this document open while reviewing, and refer to it as needed.\\
\\
Evaluation Criteria:
\\
Fluency (1-3): the quality of the caption in terms of grammar, spelling, punctuation, word choice, and sentence structure.
\\
- 1: Poor. The caption has many errors that make it hard to understand or sound unnatural. \\
- 2: Fair. The caption has some errors that affect the clarity or smoothness of the text, but the main points are still comprehensible. \\
- 3: Good. The caption has few or no errors and is easy to read and follow.
\\
\\
Example:
\\
Caption:
\\
{{Caption}}
\\
\\
Evaluation Form (scores ONLY):
\\
- Fluency (1-3):
\end{tcolorbox}
\caption{The prompt for evaluating semantic integrity (\textit{i.e.}, fluency) of generated model captions using GPT-4.}
\label{app:prompt_gpteval}
\end{table*}

\end{document}